\documentclass[sigconf]{acmart}
\usepackage{multirow}
\usepackage{hyperref}
\usepackage{xcolor}

\usepackage{tikz}
\usetikzlibrary{arrows,bayesnet,backgrounds}
\usepackage{color}
\usepackage{graphicx}
\usepackage{caption}
\usepackage{subcaption}
\usepackage{diagbox}
\usepackage{algorithm}
\usepackage{algpseudocode}

\graphicspath{{figs/}}

\newcommand\R{\mathbb{R}}

\AtBeginDocument{%
  }

\copyrightyear{2024}
\acmYear{2024}
\setcopyright{acmlicensed}\acmConference[KDD '24]{Proceedings of the 30th ACM SIGKDD Conference on Knowledge Discovery and Data Mining}{August 25--29, 2024}{Barcelona, Spain}
\acmBooktitle{Proceedings of the 30th ACM SIGKDD Conference on Knowledge Discovery and Data Mining (KDD '24), August 25--29, 2024, Barcelona, Spain}
\acmDOI{10.1145/3637528.3672014}
\acmISBN{979-8-4007-0490-1/24/08}

 \renewcommand\footnotetextcopyrightpermission[1]{}
\begin{document}

\title{Statistical Models of Top-$k$ Partial Orders}

\author{Amel Awadelkarim}
\email{ameloa@stanford.edu}
\orcid{0009-0009-2498-4976}
\affiliation{%
  \institution{Stanford University}
  \city{Stanford}
  \state{CA}
  \country{USA}
}
\author{Johan Ugander}
\email{jugander@stanford.edu}
\orcid{0000-0001-5655-4086}
\affiliation{%
  \institution{Stanford University}
  \city{Stanford}
  \state{CA}
  \country{USA}
}

\renewcommand{\shortauthors}{Awadelkarim \& Ugander}
\acmArticleType{Research}
\acmCodeLink{https://github.com/ameloa/partial-orders}

\begin{abstract}
In many contexts involving ranked preferences, agents submit \emph{partial} orders over available alternatives. Statistical models often treat these as marginal in the space of \emph{total} orders, but this approach overlooks information contained in the list length itself. In this work, we introduce and taxonomize approaches for jointly modeling distributions over top-$k$ partial orders and list lengths $k$, considering two classes of approaches: \emph{composite models} that view a partial order as a truncation of a total order, and \emph{augmented ranking models} that model the construction of the list as a sequence of choice decisions, including the decision to stop. For composite models, we consider three dependency structures for joint modeling of order and truncation length. For augmented ranking models, we consider different assumptions on how the stop-token choice is modeled. Using data consisting of partial rankings from San Francisco school choice and San Francisco ranked choice elections, we evaluate how well the models predict observed data and generate realistic synthetic datasets. We find that composite models, explicitly modeling length as a categorical variable, produce synthetic datasets with accurate length distributions, and an augmented model with position-dependent item utilities jointly models length and preferences in the training data best, as measured by negative log loss. Methods from this work have significant implications on the simulation and evaluation of real-world social systems that solicit ranked preferences.
\end{abstract}



\maketitle

\section{Introduction}
\label{sec:intro}
Ranked preferences serve as input to many consequential social systems. In election contexts, ranked choice voting (RCV) asks voters to express their preferences for candidates with a ranked list of available candidates, and these rankings are aggregated to select a winner. In school choice contexts, families express preferences for their child's education by submitting a ranked list of available program offerings to their district, and these preferences are considered when matching students to programs.

In the vast majority of such contexts, participants often submit partial rankings, specifically top-$k$ rankings. For example, a voter given a slate of $m$ alternatives
may only express strict preferences for their first $k<m$ alternatives, and not bother to provide a strict ordering of the rest.
In many preference elicitation contexts, truncating rankings in this manner can be highly consequential. For example, in San Francisco school choice, families often submit truncated rankings over only $k$ programs, but around 11\% of households do not get assigned to any of their chosen alternatives \cite{SFUSD2023}. Submitting longer rankings would only increase their placement chances, but the decision to truncate a preference list can be easily understood in terms of high search costs, a misunderstanding of the mechanism, or inflated subjective placement chances \cite{Robertson2021, Arteaga2021, Kasman2019, Lee2017a,Lee2017b,Hastings2008}. In school choice, the decision to truncate can thus mean the difference between gaining admittance to a preferred school and being relegated to participating in a later round of assignment. As a lesser but still meaningful consequence, in ranked choice voting, truncation can mean the difference between having a say in later rounds of instant run-off voting vs.\ having one's ballot ``exhausted.''

Statistical models of ranking data are often used by researchers to model demand, understand trends in preferences, or simulate counterfactual outcomes \cite{McFadden1974, Fuller1982, Train1987, Agarwal2018, Pathak2020, Laverde2022}. 
In counterfactual simulations, inquiries center around what would happen (e.g., to overall welfare) if people's alternatives or preferences changed in some structured way: some new candidate or school is added or removed from the slate of alternatives, the school-age population changes, or school prioritization changes. But investigating these counterfactuals invariably involves strong assumptions about preference lengths: either that lengths are directly taken from observed data or otherwise chosen independent of the ordered preferences themselves \cite{Pathak2020,Laverde2022}.

In this work, we argue that the problem of suitably modeling length in top-$k$ partial orders is an overlooked and highly consequential component of real-world deployments of ranking data models. 
We consider two distinct approaches to modelling top-$k$ orders: {\it composite models} and {\it augmented ranking models}. First, 
consider the probability of a partial order $Q$ of length $k$. The probability of the order can be written as the probability of its length times the marginal probability of total orders whose first $k$ elements match $Q$. Informally (see Section~\ref{sec:models} for a formal treatment), we have
\[
\Pr(Q) = 
\Pr(k)\cdot \sum_{R\in \bar{Q}}\Pr(R|k),
\]
where $\bar{Q}$ denotes the set of total orders of $m$ items whose first $k$ elements are $Q$, and $R$ is an element of that set.
Decoupling these two component probabilities, a length model and a conditional ranking model can indeed be combined to form a model of partial rankings, a model class we call \emph{composite models}. 
Under an independence assumption, such a composite model is simply:
\[
\Pr(Q) = \Pr(k)\cdot \sum_{R\in \bar{Q}}\Pr(R),
\]
dropping the conditional dependence in the ranking model.
Moving beyond independence, we consider situations where the component models (of length, of order) can be have structured dependence, and the models themselves be simple or complex.  

Beyond composite models, we also take an alternative approach that we call {\it augmented ranking}. In this approach, we augment the choice universe with an \texttt{END} token representing end-of-list, and consider a partial order as arising from a sequential choice process that models termination itself as one of the available alternatives. The agent selects $k$ alternatives and then ``chooses'' the \texttt{END} token at position $k+1$. This approach extends earlier ideas on modeling non-choice/non-purchase from choice settings to the ranking setting. Using a choice-based approach to ranking, as developed in the theory of $L$-decomposable ranking distributions \cite{Critchlow1991,Luce1959}, the probability of a partial ranking $Q$ becomes
\begin{align*}
\Pr(Q) &= \Pr(q_1\succ q_2\succ...\succ q_k) \\
&= \Pr(q_1)\cdot \Pr(q_2|q_1)\cdot...\cdot\Pr(q_k|q_1,...,q_{k-1})\cdot \Pr(\texttt{END}|q_1,...,q_k).    
\end{align*}
This lens \emph{implicitly} models list length by augmenting the choice space to include a token alternative representing end-of-list. This token can be modeled as having fixed or position-dependent selection probabilities, and we evaluate such modelling decisions in our work.

To model general dependence between ranking and length, we utilize two model stratification techniques \cite{Awadelkarim2023}. In the composite case, the ranking model can be made dependent on the length by learning $K$ separate ranking models that cover $K$ disjoint subsets of the space of partial orders, partitioned by length. The probability of a particular partial order $Q$ would then be the probability of its length, $k$, times the probability of the ordering $Q$ under the corresponding ranking model. In the case of augmented ranking models, the \texttt{END} alternative (and/or the other choice alternatives) are given $K$ distinct choice probabilities depending on its choice position within the ranking. This way, the probability of choosing an alternative, including the \texttt{END} token, is allowed to vary down the ranking. 

The primary contribution of our work is the development and evaluation of the composite and augmented ranking models, two classes of models tailored to this consequential task.
We consider several instances and sub-classes of such models, and evaluate their performances across a range of datasets with considerable variation in the size of their choice universes, $m$. 
We find that the composite approaches, which model list length explicitly, can produce more accurate sampled length distributions when generating synthetic datasets from the models, and that model stratification---a strategy we employ in both composite and augmented models---improves simulated demand over alternatives. 
While our results do not promote a universally dominant model nor model class, this work begins to formalize and taxonomize approaches for modeling distributions over the space of partial orders, and the results showcase how these components can lead to improved demand modeling and more realistic synthetic datasets.

In Section~\ref{sec:related}, we survey the related literature. 
In Section~\ref{sec:prelim}, we present notation and definitions relevant to statistical models of ranking. 
In Sections~\ref{sec:composite} and \ref{sec:augmented}, we introduce and define the two classes of partial order models discussed in this work, composite models and augmented ranking models, respectively. 
Model selection and estimation is discussed in Section~\ref{sec:selection}. 
Datasets and experiments are presented in Section~\ref{sec:results}. 
Section~\ref{sec:conclusion} concludes.

\section{Related work}
\label{sec:related}
Traditional statistical models for rankings focus on modeling total orders. 
Notable among these are variations of the Plackett--Luce (PL) model, which has been extensively used in economics, marketing, and revenue management, to learn consumer preferences from choice and rank data.
The Plackett--Luce model is a multi-stage model whereby rankings are broken into sequential choices, which are then modeled by a random utility choice model, the multinomial logit (MNL) model \cite{Train2009}.

Several works apply these models to learn preferences from partial order datasets for the purposes of rank aggregation or explanation of user preferences. For example, Zhao and Xia \cite{Zhao2019} develop theory around the task of learning mixtures of Plackett--Luce models from partial order data. In a school choice setting, Abdulkadiroğlu et al.~\cite{Pathak2020} learn an MNL choice model from partial order data to evaluate how much parents value certain school attributes. Li et al.~\cite{Li2022} develop Bayesian techniques to learn from total and partial ranking data with heterogeneous ranker preferences and item covariates.
While these works enlist the Plackett--Luce model which produces a distribution over the space of total rankings, the present work seeks to develop models that produce distributions over a larger sample space, namely the set of all partial orderings (which subsumes the former).

At other times, ranking models are specifically used to simulate real agent behavior, \emph{including} the submission of partial orders in some preference elicitation contexts. In these instances, researchers tend to learn models of total orders like the Plackett--Luce, simulate sequences of choices, and then make some assumption about the truncation of the data to achieve a dataset of partial orders. For example, Pathak and Shi~\cite{Pathak2020} learn an MNL choice model from partial order data and then simulate partial orders by manually truncating total orders to a predetermined length of ten. Laverde~\cite{Laverde2022} models partial orders to simulate counterfactual outcomes for Black and Hispanic students in Boston, but chooses to impose the original list lengths from the data.
As a significant improvement, the models developed in this work jointly model preferences and order lengths from data, and we show this model enrichment is consequential for counterfactual simulations. 

Our augmented ranking model modifies choice-based ranking to incorporate an alternative in the choice universe representing end-of-list. The idea has roots in demand modeling and revenue management for modeling the value of outside options or non-choice/non-purchase behavior \cite{kohn1974empirical, Nath2020, Gallego2015, Choo2006}. 
To our knowledge, the modelling of non-choice has not been previously employed in the ranking context for models of partial orders.

The problem of modeling partial orders is related to the problem of selecting a subset of a choice universe, as is studied in marketing, operations research, information retrieval and data mining. Such works focus on predicting or generating subsets of size $k>1$ of a universe set of $m$ alternatives, but not on ordering these subsets to yield rankings~\cite{regenwetter1998choosing, Letham2013, Benson2018SS, Benson2018SoS}.
A notable subset selection model is that of 
Regenwetter et al.~\cite{regenwetter1998choosing}. In that work, their ``subset model'' is identical to our independent composite model, but the work is (1) interested in modeling the total ordering, not modeling the subsets themselves, and (2) does not consider any dependence between size and preferences as we do in this work.

Finally, we apply model stratification in this work, a common technique in machine learning for understanding heterogeneity between different demographic and other characteristic groups in the population \cite{Kernan1999,Lash1986,Lanska2010}. Several preference modeling works use model stratification to develop tailored models for various demographic groups of interest \cite{Laverde2022, Hastings2008}. Here, we apply \emph{regularized} stratification \cite{Tuck2021} as a modeling tool for encoding dependence (in the composite case) or enriching the expressivity of a ranking model (in the augmented case), even within the same populations. Most recently, Awadelkarim et al. utilized a similar rank-based stratification technique to model heterogeneity in school choice preferences within the rankings of single individuals \cite{Awadelkarim2023}. We adopt and expand upon their idea in the present work.

\section{Preliminaries}
\label{sec:prelim}
Rankings, or ordered preferences, are seemingly intuitive objects that we engage with everyday. 
We routinely express ordered preferences over where we want to eat, who we want to lead an organization, or which movie we would like to watch with friends.
In Section~\ref{sec:rankings}, we define the notions of total and partial orders, and their related outcome spaces. 
We define statistical models, or distributions over these spaces, in Section~\ref{sec:models}. 

\subsection{Rankings}
\label{sec:rankings}
Given a collection of $m$ alternatives, $\mathcal{A} = \{a_1, ..., a_m\}$, let $\mathcal{L}(\mathcal{A})$ denote\footnote{There is a bijection between the elements of $\mathcal{L}(\mathcal{A})$ and the symmetric group, $S_m$. However, we need not invoke the group properties and operations of $S_m$, and as such, refer to the space in this work as $\mathcal{L}(\mathcal{A})$.} the set of complete rankings, or total orders, of the elements of
$\mathcal{A}$,
which contains $|\mathcal{L}(\mathcal{A})| = m!$ elements.
As an illustration, for a collection $\mathcal{A}=\{a,b,c\}$ of $m=3$ items, we have
\[
\mathcal{L}(\{a,b,c\}) = \left\{\begin{matrix}
a\succ b\succ c & b\succ a\succ c &  c\succ a\succ b\\
a\succ c\succ b & b\succ c\succ a &  c\succ b\succ a
\end{matrix}\right\}.
\]
Here the element $R=a\succ b\succ c$ is the event that $a$ is preferred to $b$ and $b$ is preferred to $c$.

Borrowing nomenclature of Xia~\cite{Xia2019} and Zhao and Xia~\cite{Zhao2019}, we focus in this work on \emph{top-$k$} partial orders, 
$Q=q_1\succ q_2 \succ ... \succ q_k[\succ \text{others}]$. 
In this case, an agent orders their top $k$ most-preferred elements of $\mathcal{A}$ and leaves the rest unordered. 
We denote by $\Omega_k(\mathcal{A})$ the set of all top-$k$ partial orders of $\mathcal{A}$, and $\Omega(\mathcal{A})$
to be the union of these sets, for any $k$.
The size of $\Omega(\mathcal{A})$ is
$$|\Omega(\mathcal{A})| = \sum_{i=1}^{m} |\Omega_i(\mathcal{A})| = \sum_{i=1}^m \frac{m!}{(m-i)!}.$$
A total ordering is a special case of a top-$k$ partial ordering where $k=m$, so we have $\mathcal{L}(\mathcal{A}) = \Omega_m(\mathcal{A}) \subseteq \Omega(\mathcal{A})$. For the collection $\mathcal{A}=\{a,b,c\}$ as above, we then have the space of top-$2$ partial orders and the (full) space of partial orders,
\begin{align*}
\Omega_2(\{a,b,c\}) &= \left\{\begin{matrix}
a\succ b & b\succ a &  c\succ a\\
a\succ c & b\succ c &  c\succ b
\end{matrix}\right\},\\
\Omega(\{a,b,c\}) &= \left\{\begin{matrix}
a & b & c\\
a\succ b & b\succ a &  c\succ a\\
a\succ c & b\succ c &  c\succ b\\
a\succ b\succ c & b\succ a\succ c &  c\succ a\succ b\\
a\succ c\succ b & b\succ c\succ a &  c\succ b\succ a
\end{matrix}\right\}.
\end{align*}
Note that the top-2 partial order $a\succ b$ and the total order $a\succ b\succ c$ represent the same overall preference profile over the items $\mathcal{A}=\{a,b,c\}$, and as such would be equal-probability events under a ranking model. These two preferences would also figure identically in many uses of ranked preferences, e.g., ranked choice voting. But the models in this work assign these two elements different probabilities on purpose, and we emphasize that this is not a bug, but a feature. The event of reporting a list of length two is different than that of listing all three and the distinction is not informationless.

Given a $k$-length partial order $Q\in\Omega_k(\mathcal{A})$, we define by $\text{Ext}(Q)$ the set of \emph{completions} of $Q$ in $\mathcal{L}(\mathcal{A})$. Specifically, let $R_{k}$ be shorthand for the first $k$ elements of $R\in\mathcal{L}(\mathcal{A})$: $R_k = r_1\succ r_2 \succ ... \succ r_k$. We have 
\begin{equation}
\text{Ext}(Q) = \{R\in \mathcal{L}(\mathcal{A}) ~|~ R_{k}=Q\}.
\label{eq:ext}
\end{equation}
For $\mathcal{A}=\{a,b,c\}$ and $Q=a$, 
the completions of this partial order are given by
\[
\text{Ext}(a) = \left\{\begin{matrix}
a\succ b\succ c & a\succ c\succ b
\end{matrix}
\right\}.
\]
Finally, for any $Q\in\Omega(\mathcal{A})$, let $k_Q$ be shorthand for the length of the partial order, $k_Q=|Q|$.
\subsection{Statistical models}
\label{sec:models}
A statistical model defines a probability distribution over a sample space $\mathcal{S}$. Specifically, let $\pi_\theta$ be that distribution, parameterized by $\theta$, such that $\pi_{\theta}:\mathcal{S}\mapsto[0,1]$. The distribution $\pi_\theta$ maps events in $\mathcal{S}$ to probabilities and therefore must sum to 1, $\sum_{s\in\mathcal{S}}\pi_\theta(s)=1$. Given a distribution over a sample space $\pi_\theta$ and a subset of its sample space $\mathcal{C}\subseteq\mathcal{S}$, we slightly overload notation and denote the probability of $C$ under $\pi_\theta$ as the sum of its constituent probabilities:
\begin{equation}
\pi_\theta(\mathcal{C}) \coloneqq \sum_{c\in\mathcal{C}} \pi_\theta(c).    
\label{eq:set}
\end{equation}
This expression holds mathematically as the events in $\mathcal{C}$ are disjoint, so the probability of their union is the sum of their individual probabilities.

In this work, we are concerned with the task of modeling distributions over the space of top-$k$ partial orders, $\mathcal{S} = \Omega(\mathcal{A})$. One modeling approach would be to assign a distinct probability to every element in $\Omega(\mathcal{A})$, learning $|\Omega(\mathcal{A})| = \sum_{i=1}^m m!/(m-i)!$ model parameters, but this strategy quickly becomes intractable as $m$ grows. 

Another strategy commonly employed for learning models from partial order data is to treat these events as marginal in the space of total orders. That is, given a distribution over linear orders, $\pi_\theta : \mathcal{L}(\mathcal{A}) \mapsto [0,1]$, consider a distribution over partial orders $\pi'_\theta: \Omega(\mathcal{A}) \mapsto [0,1]$ that is defined as 
\begin{equation}
\pi'_{\theta}(Q) \coloneqq 
\pi_{\theta}(\text{Ext}(Q)) = 
\sum_{R\in \text{Ext}(Q)} \pi_{\theta}(R).
\label{eq:piQ}
\end{equation}
This perspective alone does not allow a researcher to sample partial orders from the space, since the list length itself is taken as exogenous and left unmodeled.
In the following two sections, we present our two classes of models for efficiently parameterizing distributions over $\Omega(\mathcal{A})$.

\section{Composite models}
\label{sec:composite}
\begin{figure}[t]
\centering
\begin{subfigure}[b]{0.32\columnwidth}
\centering
\resizebox{0.65\linewidth}{!}{
    \begin{tikzpicture}
     \node[obs] (Q) {$Q$};%
     \node[latent, above=of Q, xshift=-1cm] (k) {$k$}; %
     \node[latent,above=of Q, xshift=1cm] (R) {$R$}; %
     \edge {k,R} {Q}
    \end{tikzpicture}
}
\caption{Independent}
\end{subfigure}%
\hfill
\begin{subfigure}[b]{0.32\columnwidth}
\centering
\resizebox{0.65\linewidth}{!}{
    \begin{tikzpicture}
     \node[obs] (Q) {$Q$};%
     \node[latent, above=of Q, xshift=-1cm] (k) {$k$}; %
     \node[latent,above=of Q, xshift=1cm] (R) {$R$}; %
     \node[obs, above=of k, xshift=1cm] (X) {$X$}; %
     \edge {k,R} {Q}
     \edge {X} {k,R}
    \end{tikzpicture}
}
\caption{Conditionally\\independent}
\end{subfigure}%
\hfill
\begin{subfigure}[b]{0.32\columnwidth}
\centering
\resizebox{0.65\linewidth}{!}{
    \begin{tikzpicture}
     \node[obs] (Q) {$Q$};%
     \node[latent, above=of Q, xshift=1cm] (R) {$R$}; %
     \node[latent, above=of Q, xshift=-1cm, yshift=1cm] (k) {$k$}; %
     \edge {k,R} {Q}
     \edge {k} {R}
    \end{tikzpicture}
}
\caption{Length-dependent}
\end{subfigure}%
\caption{Three dependence structures for composite models. $k\in[m]$ is a random variable representing length, $R\in\mathcal{L}(\mathcal{A})$ is a random variable representing a total order, $X\in\mathbb{R}^d$ are covariates, and $Q\in\Omega(\mathcal{A})$ is a top-$k$ partial order. Observed quantities are shaded in grey.}
\label{fig:dependencies}
\end{figure}
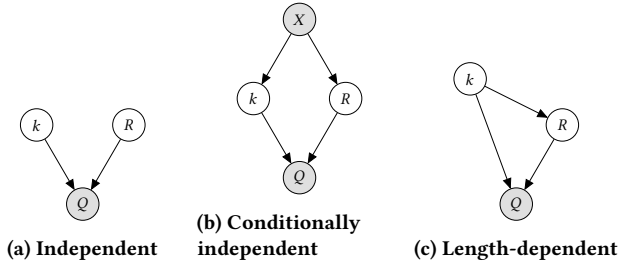
Recall our illustrative expression of the composite model from the introduction.
Now we formally define the two component models, one of length and one of preferences.
To signify these different model types, we use a superscript of $k$ to indicate a length distribution and a superscript of $R$ to indicate a ranking distribution,
\[
\pi^{k}_\theta: [m]\mapsto [0,1], \ \ \ \ \ \
\pi^{R}_\theta: \mathcal{L}(\mathcal{A})\mapsto [0,1].
\]
Through the composite lens, a partial order arises from the realization of these two component variables---a length and an order---where the order is then truncated to the sampled length to generate a partial order $Q \in\Omega(\mathcal{A})$.

We now consider both 
\emph{independent} or \emph{conditionally} independent variations of the composite model, given agent covariates, graphically modeled in Fig.~\ref{fig:dependencies}(a)-(b).
Namely, given a realization of covariates $X$, independence assumes that the distribution is decomposable in one of the following ways, with or without agent and item covariates:
\begin{alignat}{3}
\textbf{C-I:}& \quad& \pi_\theta(Q) &= \pi^{k}_\theta(|Q|)\cdot\pi^{R}_\theta(\text{Ext}(Q)),\label{eq:indep} \\
\textbf{C-CI:}& \quad& \pi_\theta(Q|x) &= \pi^{k}_\theta(k_Q|x)\cdot \pi^{R}_{\theta}(\text{Ext}(Q)|x).\label{eq:condit_indep}
\end{alignat}
where by Eq.~\eqref{eq:set}, we have $\pi^{R}_\theta(\text{Ext}(Q)) = \sum_{R\in\text{Ext}(Q)} \pi_\theta^R(R)$. We refer to Eq.~\eqref{eq:indep} and Eq.~\eqref{eq:condit_indep} as the \textbf{independent} (C-I) and \textbf{conditionally-independent} (C-CI) \textbf{composite models}.
In a recent preference modeling work where sampling partial orders was required, Laverde \cite{Laverde2022} effectively used an independent composite model, which serves as one of two baselines in this work for the task of modeling top-$k$ partial orders.

Beyond independence, we also consider length-dependence, as depicted in Fig.~\ref{fig:dependencies}(c), where the ranking is dependent on the length. That composite distribution is modelled as:
\begin{equation}
\textbf{C-LD:}\quad \pi_{\theta}(Q) = \pi^{k}_{\theta}(k_Q)\cdot\pi^{R}_{\theta}(\text{Ext}(Q)|k_Q).
\label{eq:dep}
\end{equation}
This assumption posits that \emph{how many} elements an agent chooses to rank provides additional signal as to \emph{what} they may choose to rank. 
In the school choice setting, this modeling decision allows families who rank one school program to have a different utility structure over the available alternatives than a household that ranks, say, five alternatives. Eq.~\eqref{eq:dep} is the \textbf{length-dependent} (C-LD) \textbf{composite model}. 

It is logical to ask at this point: what about a composite model where the length $k$ is dependent on the ranking $R$? While mathematically coherent, conditioning a simple length distribution on a total order is rather unwieldy. When there are many items, it requires spelling out a total order $R$ before finding that the length distribution, conditional on $R$, may actually be concentrated on very short lengths. Rather than condition the length on a total order, our augmented ranking approach in this next section can be thought of as a nuanced approach to this type of dependence, through a choice-based perspective on ranking. As an intuition for the augmented ranking model to come in Section~\ref{sec:augmented}, it effectively conditions the probability of different length outcomes upon the sequential ``choices'' made in the assembly of a ranking, up to a given truncation length. 

\subsection{Model specification}
\label{sec:composite_specification}
\subsubsection{Ranking model}
We enlist the classic Plackett--Luce (PL) model \cite{Plackett1968, Luce1977} as the ranking model, $\pi^R_\theta$, of the composite class. Under PL, we model the probability of a ranking as the product of sequential choice probabilities from shrinking choice sets. Partial orders are modeled as marginal events in $\mathcal{L}(\mathcal{A})$ as given by Eq.~\eqref{eq:piQ}. Specifically, the probability of a partial order $Q=q_1\succ...\succ q_k$ is
\begin{align}
\pi^R_\theta(Q) &= \pi^R_\theta(\text{Ext}(Q))\\
&=\prod_{j=1}^{k} \frac{\exp(\theta_{q_j})}{\sum_{a\in\mathcal{A}\setminus\{q_1,...,q_{j-1}\}} \exp(\theta_a)}. \label{eq:plQ}
\end{align}
The parameters $\theta\in\mathbb{R}^{m}$ are directly interpretable as latent item utilities. However, given features $x_{ij}\in\mathbb{R}^{d}$ of agent $i$ and item $j$, we can further model the utilities as linear in these attributes, parameterizing item $j$'s utility to agent $i$ as $\delta_j + \beta^T x_{ij},$ where $\theta=(\delta, \beta)\in\mathbb{R}^m\times\mathbb{R}^d$ are the model parameters. The probability of agent $i$ producing partial order $Q$ is then
\[
\pi^R_\theta(Q; X_i) = \prod_{j=1}^{k} \frac{\exp(\delta_j + \beta^T x_{ij})}{\sum_{a\in\mathcal{A}\setminus\{q_1,...,q_{j-1}\}} \exp(\delta_a + \beta^T x_{ia})}.
\]
When covariates are available, we enlist the linear model version of Plackett--Luce. If not covariates are available, we use the simpler model with only fixed effects $\delta_j$. 

We choose to model the length-dependence of C-LD by stratifying the preference model by list length. In this case, we use stratification to train separate preference models for people who ranked exactly one item, people who ranked exactly two items, and so on, all the way up to those who ranked exactly $K-1$ items. The final strata we reserve for those who ranked $K$ or more items.

Taking the number of strata to be $K$, a stratified ranking model is then the composition of $K$ sub-models, each with its own parameters: $\theta = \{\theta_1,...,\theta_K\}\in \mathbb{R}^{m\times K}$.
Specifically, the likelihood of a partial order $Q$ is modeled as
\[
\pi^R_{\theta}(Q | k_Q) = \pi^R_{\theta_{k'}}(Q).
\]
where $k'=\min(k_Q,K)$ and $k_Q$ is the length of $Q$; orders of length $k\in[1,K]$ are modeled by Eq.~\eqref{eq:plQ} with parameters $\theta_k$, and rankings of length $k>K$ are modeled with parameters $\theta_K$.

A possible concern with this approach is that we end up with considerably less data for each model, compared to estimating a single common model. The thinning of the training data can lead to over-parameterization which leads to overfitting. To address these concerns, we encourage models of neighboring lengths to be close to one another via Laplacian regularization \cite{Tuck2021}, borrowing the predictive power of neighboring models. 
The regularization is ``Laplacian'' because the $K$ sub-models are regularized towards each other as dictated by an accompanying \emph{regularization graph} with weights governed by the graph's Laplacian. 
Length-based stratification lends itself well to a common path graph---the model of top-1 partial orders should be similar to a model of top-2, and so on---so this regularization takes on a simple form. See Section~\ref{sec:selection} for presentation of the stratified objective function. 

\subsubsection{Length model}
\label{sec:chosen_length}
The model of length within a composite model is a distribution over $\mathcal{S}=[m]$. For full generality, we first choose to model the length as a categorical variable over these $m$ discrete categories for two of the three composite models. Namely, given parameters $\theta\in\mathbb{R}^m$, we have
\[
\pi^k_\theta(k) = p_k = \frac{\exp(\theta_k)}{\sum_{i=1}^m \exp(\theta_i)}.
\]
For the conditionally-independent composite model (C-CI), we model length as a Poisson random variable. That is, given agent covariates $x$, we define the rate parameter as $\lambda(x) = \exp(\theta^T x)$, and the resulting length distribution is given by
\[
\pi^k_\theta(k; x) = \frac{\lambda(x)^k e^{-\lambda(x)}}{k!}.
\]
The domain of the Poisson model is not bounded between $[1,m]$, so we clip extreme values into this range. 
Parameters of this model are then $\theta\in\R^{d}$ where $d$ is the dimension of covariates $x$.

\section{The augmented model}
\label{sec:augmented}
\begin{table*}[t]
    \renewcommand{\arraystretch}{1.1} 
    \centering
    \caption{Summary of models; $m$ denotes the number of choice alternatives and $k_Q$ denotes the length of partial order $Q\in\Omega(\mathcal{A})$.
    }
    \begin{tabular}{l|l|l}
        Model                                       & Distribution, $\pi_{\theta}(Q)$                           & Description\\\hline
        Composite, Independent (C-I)                & $\pi^k_\theta(k_Q)\cdot\pi^R_\theta(\text{Ext}(Q))$       & Categorical length and PL ranking.\\
        Composite, Conditionally-independent (C-CI) & $\pi^k_\theta(k_Q|x)\cdot\pi^R_\theta(\text{Ext}(Q)|x)$   & Poisson length and PL ranking, both conditional on covariates.\\
        Composite, Length-dependent (C-LD)          & $\pi^k_\theta(k_Q)\cdot\pi^R_\theta(\text{Ext}(Q)|k_Q)$   & Categorical length and length-stratified PL ranking.\\
        Augmented (A)                               & $\pi^R_\theta(\text{Ext}^+(Q))$                           & PL over $m+1$ alternatives: $\theta\in\mathbb{R}^{m+1}$.\\
        Augmented, Position-dependent (A-PD)        & $\pi^R_\theta(\text{Ext}^+(Q))$                           & PL over $m+1$ alternatives, but \texttt{END} gets $m$ utilities: $\theta\in\mathbb{R}^{2m}$.\\
        Augmented, Stratified (A-S)                 & $\pi^R_\theta(\text{Ext}^+(Q))$                           & Rank-stratified PL over $m+1$ alternatives: $\theta\in\mathbb{R}^{K(m+1)}$.
    \end{tabular}
    \label{tab:models}
\end{table*}
Consider a set of alternatives, $\mathcal{A}$, augmented with an end-of-list alternative \texttt{END} as an additional alternative in the choice universe.
A similar idea has been utilized in the revenue management literature for modeling no-purchase options \cite{kohn1974empirical,Choo2006,Gallego2015,Nath2020}.
In this model, a partial order $Q\in\Omega(\mathcal{A})$ is taken to arise from the following process: let $\mathcal{A}^+$ be the universe of alternatives plus an \texttt{END} alternative representing end-of-list, $\mathcal{A}^+ = \mathcal{A}\cup\{\texttt{END}\}$. A partial ranking $Q$ of length $k=|Q|$ is viewed as the sequential selection of each item $q_i$ from a sequence of shrinking choice sets $C_i\subseteq\mathcal{A}^+$, followed by the selection of \texttt{END} at position $k+1$. 

Let $\text{Ext}^+(Q)$ represent all linear extensions of $Q$ in $\mathcal{L}(\mathcal{A}^+)$ that begin with $Q\succ\texttt{END}$. 
Namely, $$\text{Ext}^+(Q) = \{R: R\in \mathcal{L}(\mathcal{A}^+), R_{k+1}=Q\succ\texttt{END}\}.$$
For example, given $\mathcal{A}=\{a,b,c\}$ and $Q=a$, we have
\begin{align*}
    \text{Ext}(a) &= \left\{\begin{matrix}
        a\succ b\succ c & a\succ c\succ b
    \end{matrix}\right\} \\
    \text{Ext}^+(a) &= \left\{\begin{matrix}
        a\succ \texttt{END}\succ b\succ c & a\succ \texttt{END}\succ c\succ b
    \end{matrix}\right\}.
\end{align*}
Under an augmented ranking model, the probability of observing partial order $Q$ is given by
\begin{equation}
\textbf{A: }\pi_\theta(Q) = \pi^R_\theta(\text{Ext}^+(Q)). 
\label{eq:aug}
\end{equation}
Here $\pi^{R}_\theta$ represents a probability distribution over $\mathcal{L}(\mathcal{A}^+)$, since elements $R$ here order the augmented universe that includes the \texttt{END} alternative. 

Taking the Plackett--Luce model as our baseline ranking model, we evaluate three approaches to parameterizing this new choice system. Specifically, we consider various ways to parameterize the \texttt{END} alternative relative the rest in $\mathcal{A}$, yielding the (1) na\"{\i}ve, (2) position-dependent, and (3) $K$-stratified augmented models.
The (na\"{\i}ve) \textbf{augmented model} (A) assigns the \texttt{END} alternative a fixed utility, resulting in a standard Plackett--Luce over $(m+1)$ alternatives with parameters $\theta\in\mathbb{R}^{m+1}$:
\begin{align*}
\pi^{R}_\theta(Q) &= \pi^R_\theta(\text{Ext}^+(Q))\\
&=\prod_{j=1}^{k} \frac{\exp(\theta_{q_j})}{\sum_{a\in\mathcal{A}^+\setminus\{q_1,...,q_{j-1}\}} \exp(\theta_a)} \cdot \frac{\exp(\theta_{\texttt{END}})}{\sum_{a\in\mathcal{A}^+\setminus Q}\exp(\theta_a)}.
\end{align*}
Similar to the independent composite model, our naive augmented model has been implemented for the purpose of modeling end-of-list in at least one preference modeling application \cite{Nath2020} and we consider it as the second of two baselines for our modeling task.

The \textbf{position-dependent augmented model} (A-PD) endows the \texttt{END} alternative with position-dependent utilities. Specifically, the \texttt{END} token has $m$ utilities, $\gamma\in\mathbb{R}^m$, depending on how far the choice process has gotten without terminating. The $m$ ordinary alternatives have a single position-invariant utility each, $\theta\in\mathbb{R}^m$.
\begin{align*}
    \pi^R_\theta(Q) &=\prod_{j=1}^{k} \frac{\exp(\theta_{q_j})}{\exp(\gamma_{j})+\sum_{a\in\mathcal{A}\setminus\{q_1,...,q_{j-1}\}} \exp(\theta_a)} \cdot \\
    &\hspace{1.3cm}\frac{\exp(\gamma_{k+1})}{\exp(\gamma_{k+1})+\sum_{a\in\mathcal{A}\setminus Q}\exp(\theta_a)}.
\end{align*}

Finally, the \textbf{$K$-stratified augmented model} (A-S) allows all alternatives position-dependent utilities, up to position $K$, 
\begin{align*}
    \pi^R_\theta(Q) &=\prod_{j=1}^{k} \frac{\exp(\theta^{(j')}_{q_j})}{\sum_{a\in\mathcal{A}^+\setminus\{q_1,...,q_{j-1}\}} \exp(\theta^{(j')}_a)} \cdot \\
    &\hspace{1.3cm}\frac{\exp(\theta^{(k'+1)}_{\texttt{END}})}{\sum_{a\in\mathcal{A}^+\setminus Q}\exp(\theta^{(k'+1)}_a)},
\end{align*}
where $i' = \min(i,K)$. This stratification is adopted from \cite{Awadelkarim2023} and is in contrast with the length-dependent stratification of C-LD, detailed in Section~\ref{sec:composite_specification}. There, the model assigns each \emph{user} to one of $K$ PL models, depending on how long their list was. Here, each sequential \emph{choice} up to position $K$ is governed by a unique PL model over the elements of $\mathcal{A}^+$. The resulting model has parameters $\theta=(\theta^{(1)},...,\theta^{(K)})\in\mathbb{R}^{(m+1)\times K}$. 
Similarly to the length-based stratification, we also apply Laplacian regularization between the $K$ adjacent models. Rank-based stratification also lends itself well to a common path graph as regularization graph---in this case saying the model of top-choices should be similar to a model second choices, and so on. Of course, these two stratifications (by length and by rank) are not mutually exclusive and could be applied within the same ranking model using a two-dimensional grid as the regularization graph. We leave the investigation of how such multiple stratifications interact as future work. See Table~\ref{tab:models} for a summary of our six proposed methods for modeling top-$k$ partial orders.

\section{Model selection}
\label{sec:selection}
Here we present our procedure for selecting model parameters and hyperparameters.
We estimate all model parameters using $\ell_2$-regularized maximum likelihood estimation.
Given a dataset of partial orders from $n$ participants, $D=\{Q_1, ..., Q_n\}$ where $Q_i\in\Omega(\mathcal{A})$, and an optional matrix of covariates on the agents and items, $X\in\mathbb{R}^{n\times m\times d}$, model parameters $\theta$ are chosen to maximize the likelihood of the observed dataset by minimizing the regularized negative log-likelihood (NLL),
\begin{equation}
F(D;\theta) = \ell(D; \theta) + r(\theta),
\label{eq:objective}
\end{equation}
where $\ell(D;\theta)$ is the NLL loss and $r(\theta)$ is the $\ell_2$ penalty on parameters:
\begin{align*}
\ell(D;\theta) = -\frac{1}{|D|}\sum_{Q\in D}\log\left(\pi_{\theta}(Q)\right), \ \ \ \ 
r(\theta) = \lambda||\theta||^2_2.
\end{align*}
The regularization strength $\lambda$ is a hyperparameter of the model. Our proceedure for selecting $\lambda$ and other optimization details are provided in Section~\ref{sec:setup}.

The $\ell_2$ regularization also makes our model \emph{identifiable}; a statistical model is identifiable if no two distinct sets of parameters, $\theta$ and $\theta'$, produce the same probability distributions over the sample space.
The traditional PL family of ranking distributions are non-identifiable due to their shift-invariance: $\pi^{\text{PL}}_{\theta}(s) = \pi^{\text{PL}}_{\theta+c\vec{1}}(s)$ for all $s\in\mathcal{S}$ and $c\in\mathbb{R}$. 
In this case, strategies for achieving identifiability are to fix one of the parameters, constrain their sum, or to apply regularization and obtain the minimum-norm parameter estimates \cite{Xia2019}, as we have chosen to do in this work.

For stratified models, notably the length-dependent composite model (C-LD) and the stratified augmented model (A-S), we learn $K$ PL models, over $\mathcal{A}$ and $\mathcal{A}^+$ respectively, with parameters $\theta=(\theta_1, ..., \theta_K)$, regularized toward each other via Laplacian regularization. These models are trained on $K$ stratified bands of the dataset, $D=\{D_1,...,D_K\}$, but the two models stratify the data differently.
For C-LD, $D_i$ contains all partial orders of length $i$ for all $i<K$, and $D_K$ contains lists of length at least $K$. Specifically, $D_i = \{Q \,\big|\, Q\in D, |Q|=i\}$, $\forall i<K$, and $D_K = \{Q \,\big|\, Q\in D, |Q|\geq K\}$. 
For the A-S model class, for all $i<K$, $D_i$ contains all \emph{choices} made at position $i$ across all partial orders $Q\in D$, and $D_K$ contains choices made in positions $K$ onward across all partial orders. 

Laplacian regularization in both cases is defined as:
$$
r_\mathcal{L}(\theta) = \lambda_\mathcal{L} \sum_{i=2}^K||\theta_i-\theta_{i-1}||_2^2,
$$
where $r_\mathcal{L}$ is convex in $\theta$ and $\lambda_\mathcal{L}$ is a chosen Laplacian regularization strength. 
Compared to the non-stratified objective in Eq.~\eqref{eq:objective}, the regularized, stratified objective function is the sum of $K$ decoupled model losses (each with a local $\ell_2$ regularization) and the Laplacian regularization term:
\begin{equation}
F(D;\theta) = \sum_{k=1}^K \left [ \ell(D_k;\theta_k) + r(\theta_k) \right] + r_\mathcal{L}(\theta).
\label{eq:strat_loss}
\end{equation}
Both the C-LD and A-S model introduce two additional hyperparameters, the number of strata and the Laplacian regularization gain, $(K,\lambda_\mathcal{L})$. Selection of these hyperparameters is discussed in Section~\ref{sec:setup}.

\section{Experiments}
\label{sec:results}
In this section we present an evaluation of six models---three composite models (C-I, C-CI, C-LD) and three augmented models (A, A-PD, A-S)---using seven partial order datasets, summarized in Table~\ref{tab:data}. 
Our main motivation for modelling partial ranking data is to provide a principled way to simulate partial order data for the purposes of counterfactual policy simulation or demand modeling. 
In these instances, researchers tend to learn models of total orders like the Plackett-–Luce, simulate sequences of choices, and then make some (unexplored) assumption about the truncation of the data to achieve a dataset of partial orders \cite{Pathak2020, Laverde2022}. The method implemented in \cite{Laverde2022} is actually equivalent to our independent-composite model, and another work recently implemented what we consider as a (naive) augmented model \cite{Nath2020} for the same task. Of our combined six variants of the composite and augmented classes of models, these two basic models act as our working “baselines” for modeling tractable distributions over $\Omega(\mathcal{A})$. 

We structure our experiments to answer two key questions:
\begin{enumerate}
    \item How likely is observed data, in its heterogeneously truncated form, under each of the models?
    \item How well do synthetic datasets, sampled from models with parameters estimated via maximum likelihood, simulate real demand patterns?
\end{enumerate}
In Section~\ref{sec:nll}, we report on overall goodness-of-fit on held out data as measured by negative log loss, addressing (1). To address (2), in Section~\ref{sec:synthetic} we sample datasets from the models and present demand statistics as compared with ground truth preferences in the observed data,
namely in terms of length and alternative-specific demand distributions. 
Finally, in Section~\ref{sec:downstream}, we demonstrate a downstream application of our models in the school choice realm, simulating SFUSD policy assignments using preference data generated by our models and comparing assignment outcomes under the synthetic data with assignment outcomes under the true data.

\subsection{Setup}
\label{sec:setup}
\begin{table}[t]
    \renewcommand{\arraystretch}{1} 
    \centering
    \caption{Dataset summary statistics for ranked-choice voting (RCV) and school choice (SC) data. 2018 and 2019 Board of Supervisor elections correspond to San Francisco Districts 8 and 5, respectively. Here $n$ is the number of orders in the dataset, $m$ is the number of available alternatives, and $\bar k$ is the average order length.}
    \begin{tabular}{l|c|c|c|c}
        Name                        & Label  & $n$       & $m$  & $\bar{k}$\\\hline
        2018 Board of Supervisors   & RCV1   & 33,394    &   3  &2.04\\                  
        2019 District Attorney      & RCV2   & 193,492   &   4  &2.32\\                  
        2019 Board of Supervisors   & RCV3   & 23,698    &   4  &2.06\\                  
        2019 Mayor                  & RCV4   & 178,924   &   7  &2.58\\             
        2018 Mayor                  & RCV5   & 253,866   &   8  &2.52\\\hline
        2017-18 Kindergarten        & SC1    & 3,503     &  152 &9.28\\            
        2018-19 Kindergarten        & SC2    & 3,544     &  152 &9.97\\                  
    \end{tabular}
    \label{tab:data}
\end{table}
\begin{figure}[t]
    \centering
    \includegraphics[width=\columnwidth]{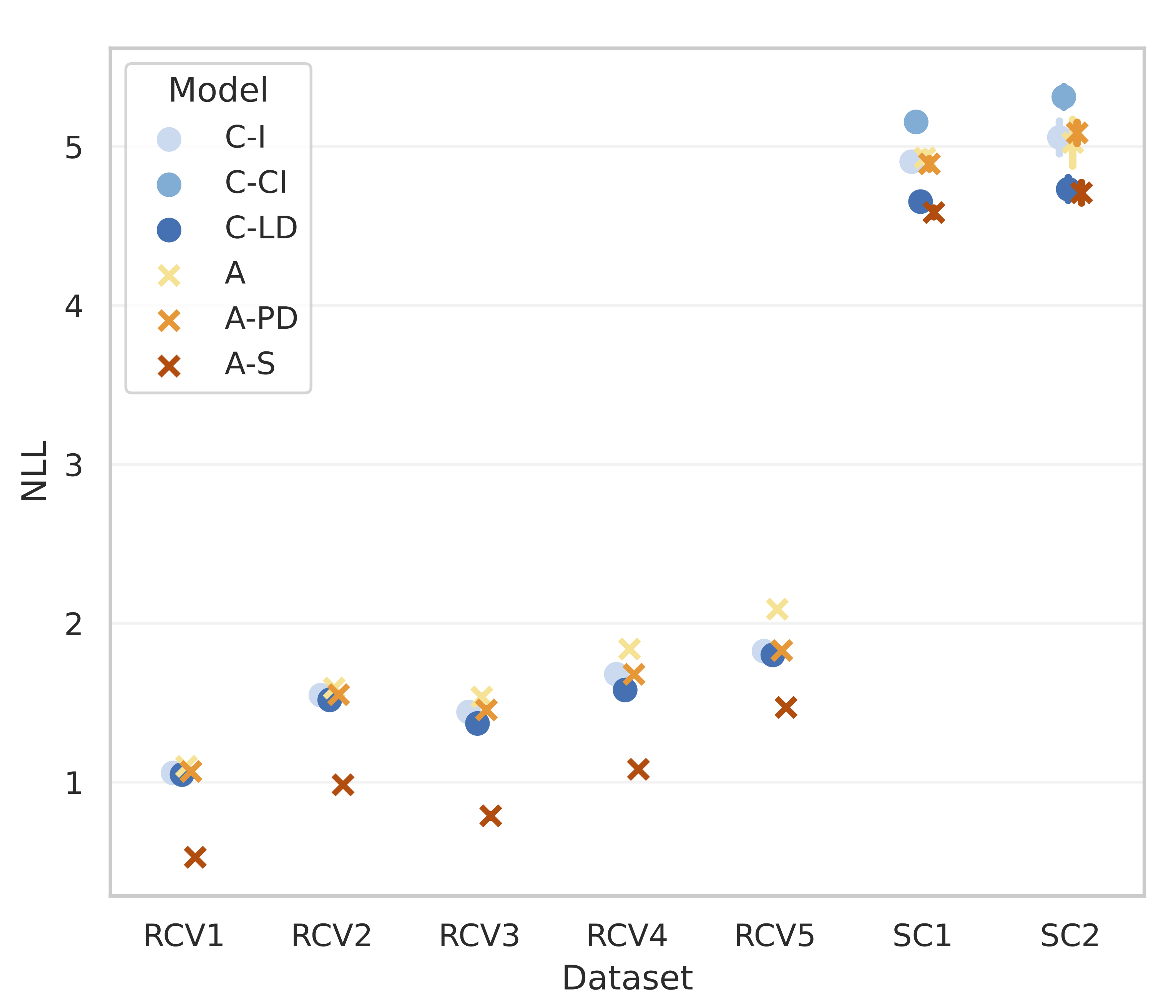}
    \caption{NLL loss (lower is better) of our test datasets under the six models in Table~\ref{tab:models}. C-I (lightest blue) and A (yellow) are the baselines.}
    \label{fig:nll}
\end{figure}

Our data comes from two sources: we use four publicly-available San Francisco ranked-choice voting (RCV) datasets from the online library of preference data, \texttt{preflib} \cite{Mattei2013}, and two (non-public) school choice (SC) datasets from the San Francisco Unified School District (SFUSD).
Both types of datasets contain strict (ie. no ties) partial orders that we treat as top-$k$ orders.
The school choice data features covariates on households and programs, which we incorporate through linear utility models.
These datasets contain PII and as such are not publicly available.
See Table~\ref{tab:data} for dataset summary statistics.

Models that enlist regularized stratification, namely the length-dependent composite (C-LD) and stratified augmented (A-S) models, introduce two additional hyperparameters, the number of stratification buckets $K$ and the regularization strength $\lambda_{\mathcal{L}}$. We conducted a grid search across suitable values of each of these parameters for each model class (composite vs.\ augmented), on a representative dataset of each type (RCV vs.\ SC). Each $(K,\lambda_{\mathcal{L}})$ pair was trained and evaluated using 5-fold cross validation on these datasets, and the values yielding the lowest average validation loss were selected. For the school choice datasets, we choose $(K,\lambda_{\mathcal{L}})=(15,1e-3)$ for the length-dependent composite model, and $(K,\lambda_{\mathcal{L}})=(15, 0)$ for the stratified augmented model. For the ranked-choice voting datasets, we choose $(K, \lambda_{\mathcal{L}})=(10,0)$ for both models. 

For model optimization, we run Adam \cite{Kingma2014}, implemented in PyTorch, with default parameters, $\text{lr} = 0.001$, $\beta = (0.9, 0.999)$, $\epsilon= 1e-8$, adding $\ell_2$ regularization with weight $\lambda= 1e-5$.
Model parameters are updated over batches of training data until reaching $\texttt{max\_epoch}=2000$ or convergence, i.e., when the absolute difference in losses is less than $\epsilon=1e-4$. 
Within each dataset, we conducted 5-fold cross-validation, training on four-fifths and evaluating on the held-out fifth. Evaluation metrics are averaged across each of these held-out portions and presented in this section. A repository containing the ranked choice voting datasets, code for our models, and a notebook for recreating the RCV results in this section can be found at \url{https://github.com/ameloa/partial-orders}.

\subsection{Goodness of fit}
\label{sec:nll}
We present the negative log likelihood (NLL) loss on the test set of our seven datasets under the six models in Figure~\ref{fig:nll}. Recall that the naive augmented (A) and independent composite (C-I) models serve as our baseline models for modeling distributions over $\Omega(\mathcal{A})$.
We see some main themes emerge from these plots. Of the augmented models, assigning only a single fixed-effect to the \texttt{END} token, as does the naïve augmented model (A), results in the worst NLL losses on observed data.
The conditionally-independent composite model (C-CI), only evaluated on the school choice datasets, yields the second worst losses; the Poisson distributional assumption on length in this model is not representative of our datasets, and thus results in poor NLL losses on observed data. 
Obviously, researchers should avoid making distributional assumptions that do not hold. 
Of the remaining four models, the stratified augmented model (A-S) demonstrates the lowest NLL loss on the RCV datasets,
and all four show similar NLL results on the school choice datasets.

\subsection{Synthetic datasets}
\label{sec:synthetic}
A main objectives of this work is to advance the available statistical models for sampling partial orders, key to generating synthetic datasets for forecasting demand or simulating counterfactual policy outcomes. 
Algorithms~\ref{alg:comp} and \ref{alg:aug} provide pseudocode of the sampling procedure used for the composite and augmented model classes, respectively. 
\begin{algorithm}
\caption{Sampling partial orders from the composite model}
\label{alg:comp}
\begin{algorithmic}[1]
\Require Learned parameters $\theta$ of a composite model.
\Ensure $Q\in\Omega(\mathcal{A})$ from composite model, $\pi_\theta$.
\State Sample length, $l\sim \pi^k_\theta$.
\State Initialize $Q=\emptyset$ and $A=\mathcal{A}$.
\For{$i=1$ to $l$}
\State Sample an alternative $a$ from $A$ according to $\pi^R_\theta$
\State $Q\gets Q \succ a$
\State $A\gets A\setminus \{a\}$
\EndFor\\
\Return $Q$
\end{algorithmic}
\end{algorithm}
\begin{algorithm}
\caption{Sampling partial orders from the augmented model}
\label{alg:aug}
\begin{algorithmic}[1]
\Require Learned parameters $\theta$ of the augmented model.
\Ensure $Q\in\Omega(\mathcal{A})$ from augmented model, $\pi_\theta$.
\State Initialize $Q=\emptyset$ and $A=\mathcal{A}^+$.
\Loop
\State Sample an alternative $a$ from $A$ according to $\pi^R_\theta$
\If{$a$ is $\texttt{END}$}
\Return $Q$
\Else
\State $Q\gets Q \succ a$
\State $A\gets A\setminus \{a\}$.
\EndIf
\EndLoop
\end{algorithmic}
\end{algorithm}

We sample $N=100$ synthetic datasets, $\tilde{D}^{(i)} = \{Q^{(i)}_1,..., Q^{(i)}_n\}$ for $i\in[N]$, from each of the six models, for each of the seven datasets in Table~\ref{tab:data}.
We use the same covariates $X\in\mathbb{R}^{n\times m\times d}$ for sampling as in training, so the synthetic datasets are seen as simulated preferences
of those $n$ households.

\begin{figure}[t]
    \centering
    \includegraphics[width=\columnwidth]{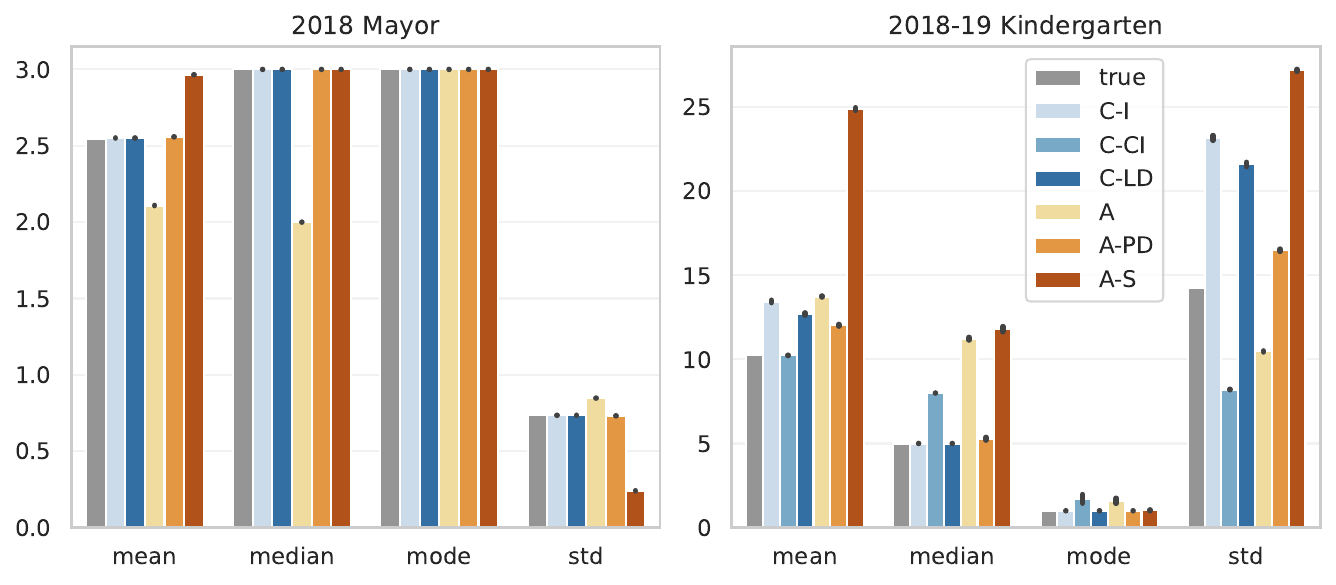}
    \caption{
    Sampled length distributions statistics on a representative RCV (left) and SC (right) dataset. Statistics of their true distributions in grey. C-CI was not evaluated on the RCV datasets as no voter covariates were available with the data.}
    \label{fig:syntheticlength}
\end{figure}

Each dataset $\tilde{D}^{(i)}$ produces an empirical length distribution over $n$ examples, and we summarize averaged statistics in Figure~\ref{fig:syntheticlength}, taking one RCV and one SC dataset as representative examples.
We see that the biggest swings in either direction --- longer or shorter lists --- come from the augmented class. The naive augmented model (yellow) samples orders that are shorter than the rest, with higher standard deviations on RCV dataset. The single \texttt{END} token of this model learns a high utility to explain shorter-than-full-length lists seen in the observed data. The stratified augmented model, where all alternatives get position-dependent utilities, produces full-length lists in the RCV case. In the school choice case it produces longer, and more dispersed lists. The latter observation suggests that if the sampled list happens to get longer than the modal length of 1, the utility of other alternatives remains higher than \texttt{END} token and the list length continues to grow. Surprisingly, the position-dependent augmented model, where \texttt{END} token gets position-dependent utilities but the other alternatives utility is fixed down the ranking, matches true lengths distributions as well as or better than when lengths are explicitly modeled as done by the composite class. Overall, the augmented class produces more varied length distributions than the composite class, but still has potential to fit lengths well. The composite class, which models list length explicitly, fits the true length distribution reliably well.

\begin{figure}[t]
    \centering
    \includegraphics[width=\columnwidth]{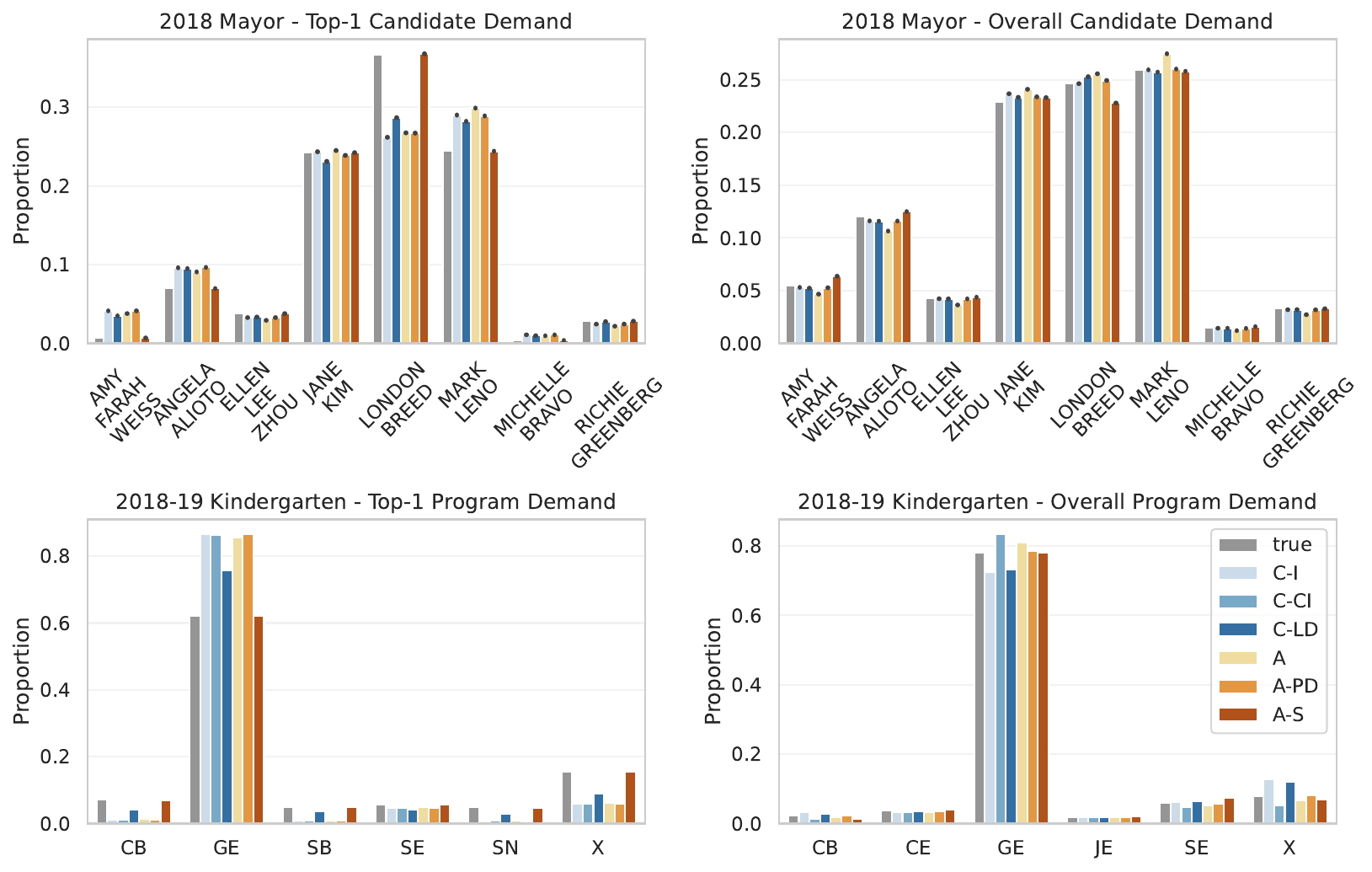}
    \caption{Synthetic demand over 2018 SF Mayoral candidates (top row) and 2018-19 SFUSD program types (bottom row). True demand in grey. Left plots show proportion of choices in first position, right plots show proportion of choices overall.}
    \label{fig:syntheticdemand}
\end{figure}

In Figure~\ref{fig:syntheticdemand} we showcase the overall and top-choice demand of each candidate in the chosen RCV dataset in the top row, and of the most popular program types in the SFUSD school choice dataset in the bottom row. In the latter plots, rather than presenting every available program, the x-axis buckets school programs by program type — General Education, Chinese Biliteracy, etc. 
In the left plots of Figure~\ref{fig:syntheticdemand}, we see that the stratified augmented model (red) fits top choice alternatives best in both the RCV and SC datasets. It is the only model to learn position-1 alternative utilities that are independent of those down rank.
The models are showcasing similar performance to one another and matching the overall distribution of demand well in the right plots.

\subsection{School assignment outcomes from simulated preferences}
\label{sec:downstream}
\begin{figure}
    \centering
    \includegraphics[width=\columnwidth]{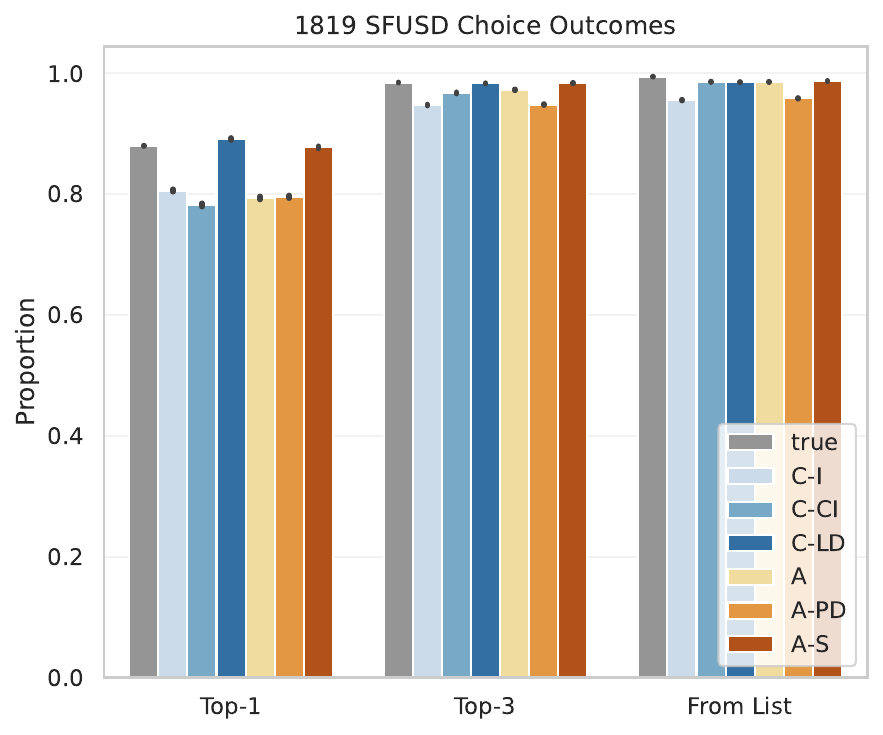}
    \caption{Assignment outcomes using synthetic school choice datasets sampled from our 6 models compared with true outcomes in grey. Proportion of students who were assigned to their top-1, a top-3, or any one of their listed alternatives (as opposed to non-assignment).}
    \label{fig:da}
\end{figure}

Our final analysis aims to demonstrate downstream simulation ability of our proposed methods. We generated $N=100$ synthetic school choice datasets and simulated real assignment outcomes using SFUSD's 2023 assignment policy. In Figure~\ref{fig:da}, we summarize assignment outcomes of our six models compared with simulated outcomes using true preference lists. Overall, we find that the stratified models, C-LD and A-S, mimic true outcomes best as they are more expressive preference models than their non-stratified counterparts. The independent composite (C-I) and position-dependent augmented (A-PD) models seem to result in datasets where fewer students gain access to their top choices than in assignments based on the real preferences. These simpler models learn universal weights over certain agent-item attributes, and may lead to ``monoculture'' effects \cite{kleinberg2021algorithmic} and greater competition for program offerings than their stratified counterparts. If C-I and A are our baselines, we see there is great potential in using some of the ideas developed in this work, esp. the more expressive stratified models.

\section{Conclusions}
\label{sec:conclusion}
In this work, we study the statistical modeling of partial orders \emph{not} as marginal events in the space of total orders, but as deliberate individual events. We developed two approaches to modeling top-$k$ partial orders with three implementations each. The sample space of these models are the space of partial orders directly, $\Omega(\mathcal{A})$, and they provide researchers with the ability to sample meaningful synthetic datasets of this type.
Under the augmented class of models, whereby end-of-list is modeled as another alternative in the universe, the new alternative should have position dependent fixed-effects, not simply one latent utility. 
The composite class, which models list length directly, generally produces synthetic datasets that exhibit the most accurate length distributions.
Model stratification, applied in both the length-dependent composite model (C-LD) and the $K$-stratified augmented model (A-S), is an important aspect of preference modeling, improving the accuracy of simulated demand and mimicking true school choice outcomes best in our downstream application experiments.

There are some known limitations with the models and analyses presented in this work. Plackett-Luce ranking models learn ranking distributions over a fixed choice universe, $\mathcal{A}$. As such, the methods presented here are not capable of simulating counterfactuals that add or change the properties of the alternatives in the item set.
They are, however, fully applicable to counterfactual evaluations that study when, e.g., the distribution of household covariates changes. Approaches to relaxing these two requirements on the alternative set would increase the range of applications of our models and is an area of future work. 
Additionally, further analysis is needed to characterize the ranking distributions that are expressible by one or both of our two methods, understand how canonical the representations of each type are, and furnish theoretical guarantees around identifiability and convergence.
\begin{acks}
    We thank Irene Lo and Itai Ashlagi for being our liaisons to the San Francisco Unified School District's preference data and Arjun Seshadri for his ongoing modeling support and advice. This work was supported by NSF CAREER Award \#2143176.
\end{acks}

\bibliographystyle{plain}
\balance
\bibliography{sample-base}

\end{document}